\documentclass{article}
\usepackage{spconf,graphicx}
\usepackage{amsmath,amssymb,amsfonts}
\usepackage{multirow}
\usepackage{bbm}
\usepackage{color}
\usepackage{algorithmic}
\usepackage{graphicx}
\usepackage{textcomp}
\usepackage{enumitem}
\setlist{nosep, leftmargin=14pt}

\usepackage{mwe} 

\title{TriAug: Out-of-Distribution Detection for Imbalanced \\Breast Lesion in Ultrasound}
\name{Yinyu Ye \qquad Shijing Chen\qquad Dong Ni \qquad Ruobing Huang$^\dag$ \thanks{\dag Corresponding author. E-mail address: ruobing.huang@szu.edu.cn}}
\address{Guangdong Key Laboratory of Biomedical Measurements and Ultrasound Imaging,\\ School of Biomedical Engineering, Shenzhen University Medical School, Shenzhen University, China}
\begin{document}
%
\maketitle
\begin{abstract}
Different diseases, such as histological subtypes of breast lesions, have severely varying incidence rates. Even trained with a substantial amount of in-distribution (ID) data, models often encounter out-of-distribution (OOD) samples belonging to unseen classes in clinical reality. To address this, we propose a novel framework built upon a long-tailed OOD detection task for breast ultrasound images. It is equipped with a triplet state augmentation (TriAug) which improves ID classification accuracy while maintaining a promising OOD detection performance. Meanwhile, we designed a balanced sphere loss to handle the class imbalanced problem. Experimental results show that the model outperforms state-of-art OOD approaches both in ID classification (F1-score=42.12\%) and OOD detection (AUROC=78.06\%).
\end{abstract}
\section{Introduction}
Breast cancer is the most common cancer worldwide, especially affecting women~\cite{2Global2021}. Ultrasound (US) is one of the common tools for determining breast abnormality, particularly for young women with dense breasts~\cite{1Global2022}. To better assist the diagnosis, many computer-aided diagnosis (CAD) systems have been proposed to identify the malignancy likelihood of lesions in US~\cite{AW3M, BeckerMueller, Moon2020}. However, both benign and malignant lesions encompass a diverse spectrum of histological subtypes, and relevant information may further inform the treatment planning~\cite{makki2015}. Directly inferring these subtypes based on US images is not only challenging for clinicians but also for machine learning algorithms as some of these subtypes have a rare incidence  (e.g., neuroendocrine carcinoma accounts for 2\% of breast cancer)~\cite{makki2015}. This may result in the omission of such subtypes throughout the entire data collection process, with either no representation or an exceedingly limited number of samples. Therefore, it may be more plausible to flag these samples during the testing to seek human intervention, rather than forcefully constructing classifiers that produce over-confident predictions. 

\begin{figure}[ht]
\centerline{\includegraphics[scale=0.09]{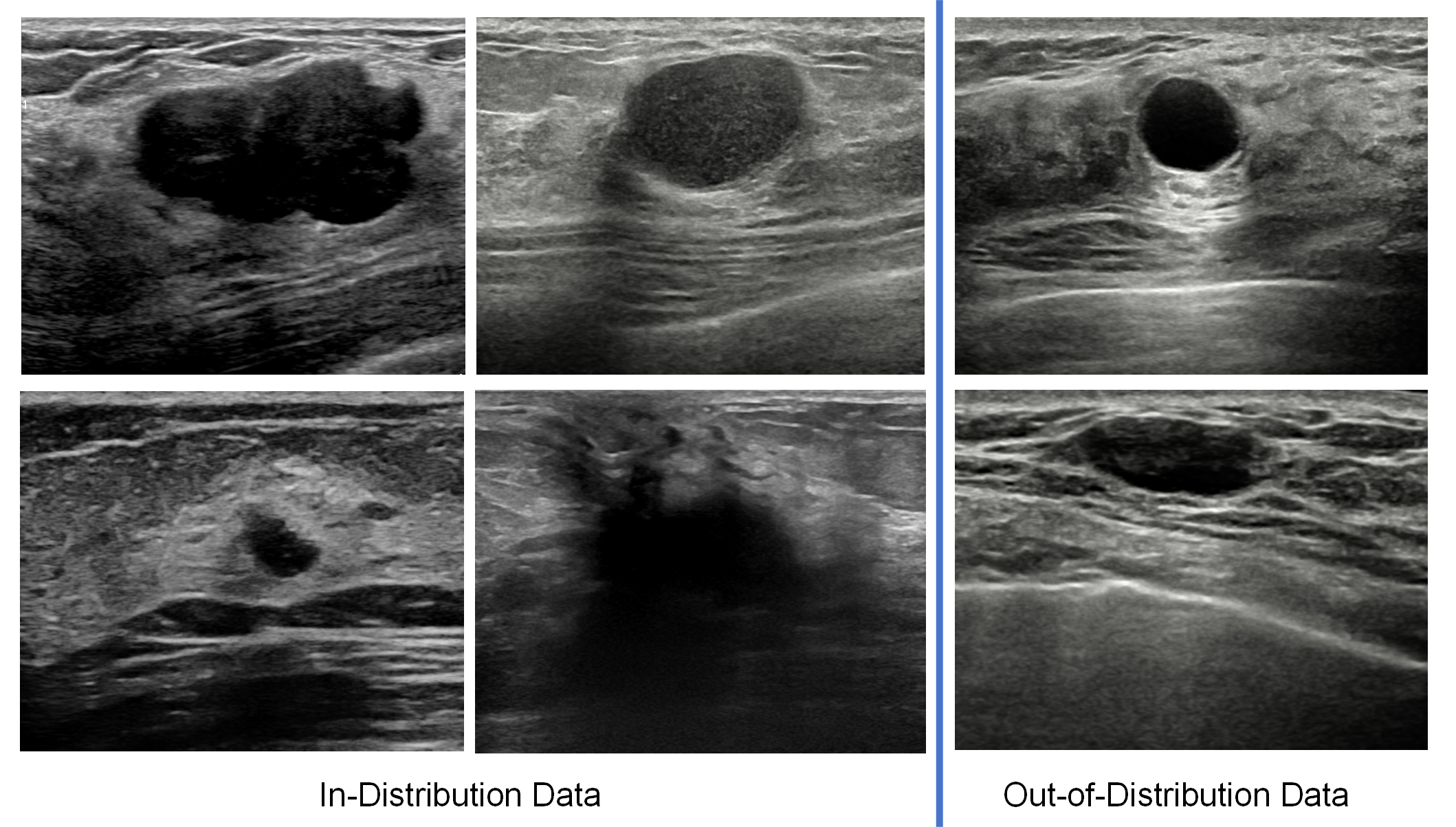}}
\caption{ID samples vs OOD samples of breast lesions in US.}
\label{fig1}
\end{figure}
This problem is closely related to an existing research topic in computer vision, known as Out-of-Distribution (OOD) detection. Naive models that only witnessed In-Distribution (ID) data during the training may classify OOD samples as one of the ID categories. It is crucial to establish the ability of models to reject OOD samples to ensure their reliability and robustness in real-world applications. The first solution to address this leveraged the maximum softmax probability as the indicator score of OOD identification~\cite{MSP}. However, the softmax classifier is prone to overconfident predictions for OOD samples. To improve this,~\cite{ODIN} used temperature scaling and input perturbation to amplify the separability between the ID and OOD samples.
\cite{MD} uses the minimum Mahalanobis distance to all class centroids for OOD detection and its core idea is that the OOD samples should be relatively far away from centroids or prototypes of ID classes.~\cite{2021Open} demonstrates the performances of the ID set and OOD set are highly correlated. Therefore, following methods aim to increase model capacity for better feature extraction.~\cite{RegMixup} utilizes mixup as a regularizer to improve ID  performance as well as OOD robustness.

Although these approaches have made a sequence of advances, the common problem settings do not resemble real-world clinical scenarios. The main differences are: 1) Their training sets are well-balanced (e.g.,  CIFAR-10, ImageNet), while medical datasets often exhibit class imbalance. Therefore, these models tend to produce biased predictions and overfit to a particular subset. 2) They commonly focus on coarse-grained features (e.g. cars vs. horses), while medical images exhibit minutiae differences that are difficult to capture (see Fig.~\ref{fig1}). 
Some efforts have been made to perform OOD detection in medical images~\cite{medood1, medood2}, while challenges are different in breast US images.

In this work, we propose the first framework for OOD detection for breast lesion subtypes in the US, an area that, to the best of our knowledge, has not been explored by others. It is capable of handling the aforementioned challenges and achieving better performance in both OOD and ID data. The main contributions are:
1) A triplet state augmentation (TriAug) based framework that increases data variability to further optimize latent space, while avoiding representation disruption or entanglement.
2) A balanced sphere loss function that utilizes both subsphere equilibrium to battle class imbalance and a prior knowledge-guided hypersphere constraint to further modify the learned embeddings.

\begin{figure*}[ht]
\centerline{\includegraphics[scale=0.156]{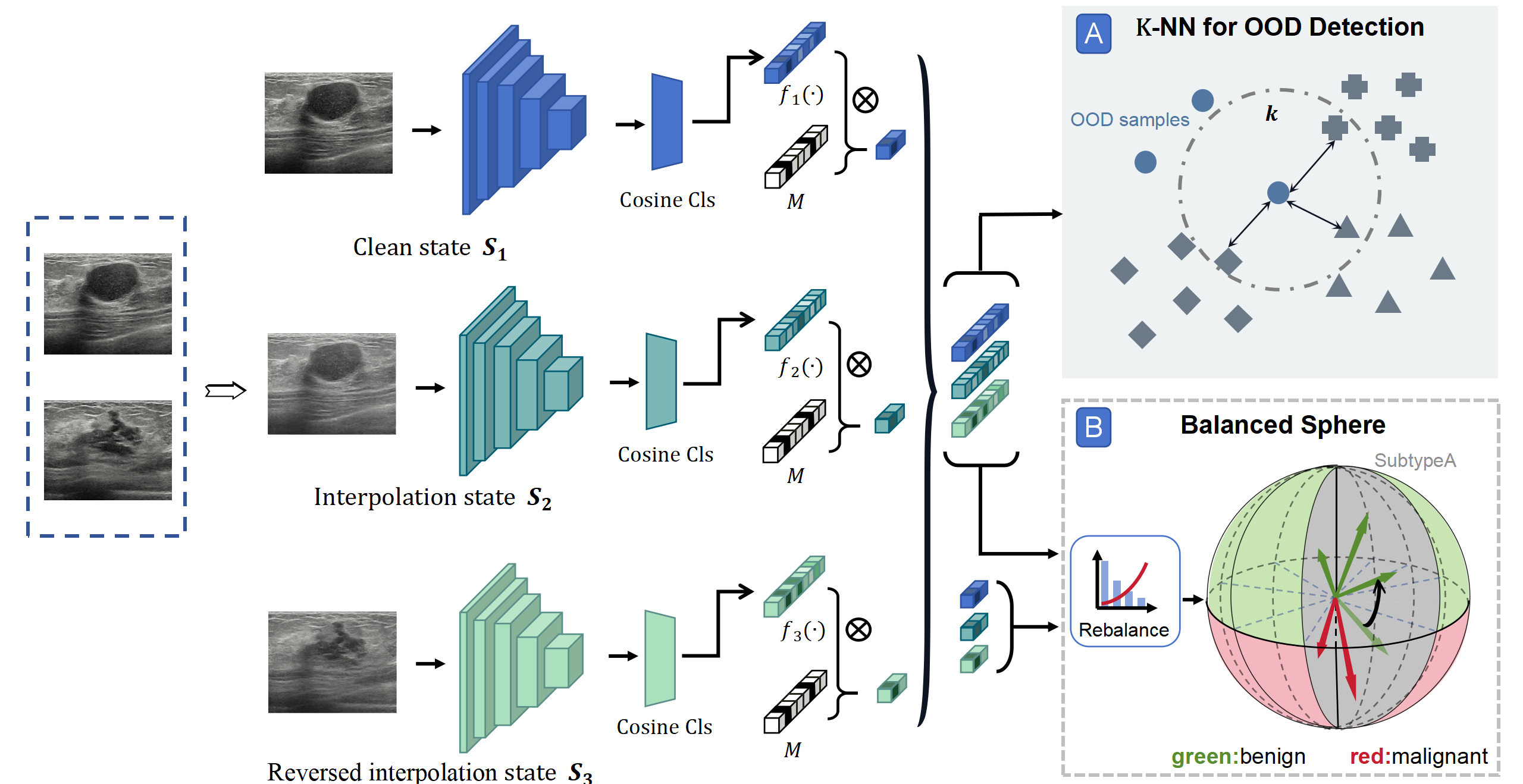}}
\caption{Schematic of the proposed framework.}
\label{fig2}
\end{figure*}

\section{Method}
A trustworthy visual recognition system should not only produce accurate predictions on known context but also correctly detect examples from unknown classes during the test. 
Formally, given a multi-class ID dataset $D_{IN}=\{(x_i, y_i)\}_{i=1}^{n}$ , where $y_i \in C $ is the class label of the sample $x_i$ and a unknown test set $D_{test}=\{(x^{test}, y^{test})\}, y^{test}\in R \varsupsetneqq C$, where $R$ is a superset of C. The total number of training data is $n=\sum_{c=1}^Cn_c$, where $n_c$ denotes the number of samples of class $c$. We define a model $f(\cdot)$ that is trained using $D_{IN}$ to map $x_i$ to an ideal high dimensional latent space: $x \rightarrow f(x_i)$ for subsequent classification $\hat{y}_i =Cls(f(x_i))$. During the test, given a test sample  $x^{test}$, the task is not only to yield the multi-class classification prediction $\hat{y}_{test}$ but also to produce a confidence score of $x^{test}$ belongs to the OOD. 
The core of distance-based methods is the testing OOD samples should be relatively far away from the centroids of  ID classes. Due to the obscurity of OOD data, we compute the $k$-Nearest Neighbor (KNN) distance between the test input embedding and the embeddings of the training set to exempt prior assumption of feature distribution.
Specifically, we use the normalized embeddings extracted from the model $z=f(x)/\parallel f(x)\parallel_2$. Denote the embedding set of the training set as $\mathbb{Z} = (z_1, z_2, ..., z_n)$,  then giving a normalized feature $z^{test}$ of  $x^{test}$, their Euclidean distance could be calculated through:
\begin{equation}
d_i=\parallel z_i-z^{test} \parallel_2,
\label{eq1}
\end{equation}
where $z_i\in \mathbb{Z}$. $\{d_1, d_2,...,d_n\}$ is then re-arranged in the ascending order :$\{d'_i\}_1^n$. The decision function for OOD detection is given by:
\begin{equation}
G(z^{test},k)=\mathbbm{1}\{-d'_k\ge \tau \},
\label{eq2}
\end{equation}
where $\mathbbm{1}\{\cdot\}$ is the indicator function. $d'_k$ is the $k$-th smallest distance and is used as the confidence score. $k$ is empirically set as 1000 in this work.
$\tau$ is the threshold that separates the OOD classes from the ID ones. Its value could be flexibly adjusted to determine the fraction of ID data (e.g.,95\%) that is correctly classified. Such design permits the existence of intra-class variation while highlighting OOD samples with embeddings that are distant from that of the ID classes. 
Given such a flexible framework, the remaining core issue is to find a robust $f(\cdot)$ that produces representative and discriminative embeddings for both ID and OOD data.
\subsection{Triplet state Augmentation}
Augmentation techniques have been widely used to improve the robustness and performance of models on ID data~\cite{RandAugment}.  For example, Mixup~\cite{mixup} helps the model training to improve generalization and combat memorization of corrupt labels.
However, later work~\cite{RegMixup} observed that Mixup's reliability degrades when exposed to completely unseen samples (e.g., OOD ones) since it tends to produce high entropy for almost all the samples it receives. To build a $f(\cdot)$  that exploits the advantage of Mixup while suppressing its confidence reducing behavior, we propose a TriAug framework (see Fig.~\ref{fig2}). It comprises triplet states $\mathbf{S}=\{\textbf{S}_1, \textbf{S}_2, \textbf{S}_3\}$ with three specific augmentation strategies. Note that each state uses the same ResNet50 as backbone and cosine classifier~\cite{cos} for simplicity, while it could be replaced for other applications. 

Specifically, a clean state $\textbf{S}_1$ adopts the RandAugmentation~\cite{RandAugment} on an individual sample and seeks to discern the original data distribution. The interpolation state $\textbf{S}_2$ uses a mixture of samples to explore a better latent representation and helps to increase the classification performance on ID data. 
The mixed inputs for $\textbf{S}_2$ is interpolated with the subsequent rule: $x_{mix} = \lambda x_i + (1-\lambda)x_j$, where $\lambda \sim \beta\{1,1\}\in [0,1]$, and the following loss function is minimized by:
\begin{equation}
\mathcal{L}_{mix} = \lambda\mathcal{L}_{bs}(x_i, y_i)+ (1-\lambda)\mathcal{L}_{bs}(x_j, y_j).
\label{eq3}
\end{equation}
The reversed interpolation state $\textbf{S}_3$ is conjugated with $\textbf{S}_2$, and further informs the framework regarding the interrelationship between the interpolated samples.
The reversed mixed inputs for $\textbf{S}_3$ can be denoted as $x_{rmix} = (1-\lambda)x_i + \lambda x_j$ and it loss function is minimized by:
\begin{equation}
\mathcal{L}_{rmix} = (1-\lambda)\mathcal{L}_{bs}(x_i, y_i)+ \lambda \mathcal{L}_{bs}(x_j, y_j).
\label{eq4}
\end{equation}
Such streamlined configuration could help to mitigate the representation disruption and avoid inter-class feature entanglement. The outputs of the three states are averaged to produce the final embeddings: 
\begin{equation}
f(x)=\frac{1}{|\mathbf{S}|}\sum_{i\in\mathbf{S}}f_i(x).
\label{eq5}
\end{equation}
Within the proposed framework, $\textbf{S}_1$ helps to stabilize the original ID learning, $\textbf{S}_2$ enhances training data variability and reduces overfitting, while $\textbf{S}_3$ reinforces inter-class dependency to restrain the high entropy caused by augmentation. 
Later experiments verify that TriAug is able to enhance the latent representation to boost the performance of both ID classification and OOD detection.

\subsection{Balanced Sphere Loss Function}
Another issue that requires consideration is the class imbalance, as the head classes will dominate the training process when handling medical long-tailed datasets. Moreover, histological subtypes of lesions share similar sonographic features and their differences are fine-grained (see Fig.~\ref{fig1}). To better overcome these, we propose a balanced sphere loss to better constrain the learned embeddings in both the subsphere level and the hypersphere level in the latent space (see  Fig.~\ref{fig2}B).
Specifically, we assume the full distribution of ID data as a perfect sphere in an ideal latent space, while each class occupies different subspheres. Intuitively, head classes would take up larger regions than that of tail classes. To enhance the representability of the latter, we propose to enforcing subshpere equilibrium by adjusting the areas of these subshperes to be isometrical via a balancing strategy~\cite{bsce} (e.g., the grey region in Fig.~\ref{fig2}B). To be precise, the embeddings are modified by $f^{'}(x) = f(x) + log\pi$, where $\pi_c=\frac{n_c}{n}$ denotes the frequency of class $c$, to compensate for the skewed distribution.
\begin{equation}
\mathcal{L}_{s} =\frac{1}{N} \sum_{i \in \textbf{S}} \sum_{n=1}^{N}-y_nlog\sigma(f^{'}_i(x_n)),
\label{eq6}
\end{equation}
where $\sigma(\cdot)$ is the softmax function.

Meanwhile, as different subtypes in ID are individually attributed to either benign or malignant neoplastic lesions, the whole sphere can be divided into two (see the green and red hemispheres in Fig.~\ref{fig2}B). This clinical prior knowledge (denoted as $y^*$) could further pull the embedding vectors of the same subtype into convergence to a specific hypersphere. To exploit this, we designed a mapping algorithm to transfer the original subsphere predictions to hypersphere ones. In specific:
\begin{equation}
f^*(x) = Concat(\Sigma (f^{'}(x)\cdot{M}),\Sigma (f^{'}(x) \cdot (1-{M}))),   
\label{eq7}
\end{equation}
\begin{equation}
{m}_c = \left\{ \begin{array}{ll}
0 & y^*=\textrm{benign}\\
1 & y^*=\textrm{malignant}
\end{array} \right.
\label{eq8}
\end{equation}
where ${M}=[m1,...,m_c,...,m_C]  $ is a binary mask with a size of $C$ to split all subtypes into benign/malignant groups. The hypersphere loss can then be constructed through:
\begin{equation}
\mathcal{L}_{h} = \frac{1}{N} \sum_{i \in \textbf{S}} \sum_{n=1}^{N}-y_n^*log\sigma(f_i^*(x_n)).
\label{eq9}
\end{equation}
The Balanced Sphere loss can be written as: $\mathcal{L}_{bs} = \mathcal{L}_{h} + \mathcal{L}_{s}$.
\begin{table*}[htbp]
    \centering
    \caption{Comparison with SOTA methods regarding both ID classification (column 2-4) and OOD detection (column 5-8). Rows 5-6 demonstrate the performance of variants of the proposed method. }
    \label{table:1}
    \scalebox{0.92}{
    \begin{tabular}{l|ccc|cccc}
     \hline
     \multirow{2}{*}{Methods} & \multicolumn{3}{c}{ID} & \multicolumn{4}{|c}{OOD} \\
     \cline{2-8}
     & F1-score (\%) $\uparrow$ & Recall (\%) $\uparrow$ & Precision (\%) $\uparrow$ & AUROC (\%) $\uparrow$ & FPR@95 (\%) $\downarrow$& AUPR-In $\uparrow$ & AUPR-Out $\uparrow$\\
    
    \hline
    Baseline\cite{MSP} & 31.48 & 27.89 & 36.14 & 72.76 & \underline{59.74} & 62.95 & 77.88\\
    
     RegMixup\cite{RegMixup} & 34.33 & 28.49 & \underline{43.20} & \underline{75.85} & \textbf{54.55} & \underline{65.75} & \underline{81.33}\\
     OLTR\cite{OLTR} & 35.67 & 42.18 & 30.90 & 59.23 & 73.91 & 52.03 & 67.39\\
     PixMix\cite{pixmix} & 36.69 & 31.71 & \textbf{43.51} & 69.68 & 71.28& 61.99 & 73.99\\
    
    \hline
   TriAug($S_1$*3)& 37.87 & 40.77 & 35.35 & 61.09 & 88.85 & 56.44 & 63.41\\
    TriAug($S_2$*3) & \underline{40.72} & \underline{47.20} & 35.80 & 56.85& 97.95& 54.68 & 54.16\\
    TriAug(Ours) & \textbf{42.12} & \textbf{47.76} & 37.66 &\textbf{78.06} & 61.54 & \textbf{72.57} & \textbf{81.34}\\
    \hline
    \end{tabular}
    }
\end{table*}
    
\section{Experiments}
\textbf{Datasets.} Our in-house dataset contains 13 classes of different histological subtypes of breast lesions. The most frequent 8 classes besides the cysts (i.e., Fibroadenoma, Fibrocystic Breast Disease, Intraductal Papilloma, Sclerosing Adenosis, Mastitis, Benign Phyllodes Tumor, Invasive Carcinoma, Ductal Carcinoma In Situ) are selected as ID classes, which includes 3905 cases. The other 5 classes were used to build the OOD test set (i.e., Cysts, Hamartoma, Neuroendocrine Carcinoma, Atypical Hyperplasia, Intracystic Papillary Carcinoma), which contains 740 cases. All ground truth labels were obtained through biopsies. Each class of the ID set is randomly split at the patient level in 7:1:2 for training, validation and testing, while the OOD set is only available during the test. The dataset is long-tailed with a maximum class imbalance ratio of 47.975:1, presenting a significant challenge.

\noindent\textbf{Experiments.} The model performance of the ID set is evaluated using precision, recall, F1-score. The standard metrics for OOD detection tasks are AUROC, FPR@95 (the FPR when TPR is at 95\%), AUPR-In (ID data as positive), AUPR-Out (OOD data as positive). We set MSP\cite{MSP} as a baseline without augmentation and compared the proposed method to front-end SOTA approaches for OOD, i.e., RegMixup~\cite{RegMixup}, Pixmix~\cite{pixmix}),and OLTR~\cite{OLTR}. The latter also targeted at the long-tailed dataset. Results are reported in Table~\ref{table:1}. Note that these methods all used \cite{MSP} as post-processing.
We also compare popular post-hoc approaches including MSP~\cite{MSP}, ODIN~\cite{ODIN}, MD~\cite{MD}(see Table~\ref{table:2}).
Another factor worth investigating is whether the proposed triplet state augmentation design is beneficial. Thus, we also implement the proposed framework with three identical $S_1$ or $S_2$. They also use a three-branch design for a fair comparison, while each branch uses the identical augmentation strategy. 

\noindent\textbf{Implementation.} All experiments were implemented in PyTorch with a GeForce RTX 3090 GPU. SGD optimizer with momentum was used with an initial learning rate setting to 1e-3, and weight decay was set to 2e-4. 

\section{Results and discussion}
Table~\ref{table:1} shows that the proposed framework achieved superior performance in both ID and OOD data than the competing methods. Compared to the baseline, RegMixup achieved higher performance in both ID and OOD data. It is interesting to see that the OLTR and Pixmix achieved relatively high performance on ID classification while obtaining a low performance on OOD detection. This may be explained that our dataset has a small inter-class difference (see the ID classes in Fig.~\ref{fig1}) and the overall data size is limited.
Row 6 of Table~\ref{table:1} displays the performance of the framework which uses interpolation state monotonically for all three states---a common strategy applied by many ensemble-based augmentation approaches. Results exhibit that they both obtained higher ID performance than Row 5, while their ability in OOD detection might be restricted.
It verifies that a strong augmentation technique could increase model robustness while disrupting the latent space and hampering OOD detection. 
The TriAug-based framework (Row 7) obtained a 21.21\% increase in AUROC and boosted the correctness of the ID set compared to Row 6. Equipped with the balanced sphere loss, it obtained the highest F1-score among all competing methods, verifying its ability to handle the long-tailed issue while ensuring the OOD detection performance.

Table~\ref{table:2} displays the results of the post-hoc methods of OOD detection. MSP and ODIN are the softmax-based methods and obtained lower performance. It may rooted from that the OOD samples share visually similar features with ID samples in our task (see Fig.~\ref{fig1}) and require stronger latent embedding to differentiate. MD obtained a higher FPR score as it assumes ideal multivariate Gaussian distributions, which may not hold for all testing scenarios. Our method achieves superior performance (more than 17.74\% increase in AUROC) as it exempts prior distributional assumption while its TriAug design produces robust and discriminative embeddings.
\begin{table}[t]
    
    \centering
    \caption{OOD detection performance of our postprocessing method and comparison methods. }
    \label{table:2}
    \scalebox{0.85}{
    \begin{tabular}{ccccc}
     \hline
    Methods & AUROC (\%) & FPR@95 (\%) & AUPR-In & AUPR-Out  \\
    \hline
     MSP\cite{MSP} & 56.04 & 95.96 & 53.45 & 54.19\\
     ODIN\cite{ODIN} & 57.96 & 95.38 & 54.06& 56.12\\
     MD\cite{MD} & 60.32 & 86.73 & 56.01& 61.75\\
     Ours & \textbf{78.06}& \textbf{61.54}  & \textbf{72.57}& \textbf{81.34}\\

    \hline
    \end{tabular}
    }
\end{table}
\section{Conclusion}
In this paper, we proposed a TriAug framework for OOD detection and multiple subtype classification for breast ultrasound images. It is equipped with a balanced sphere loss that achieved superior performance in both OOD samples and ID samples with long-tailed distribution. Due to the space limitation of the paper, this work did not discuss the impact of domain shift on OOD detection. Future studies may investigate further on this as the proposed methodology is general and could be applied to other application.


\vfill
\pagebreak

\section{Acknowledgments}
\label{sec:acknowledgments}

This study was supported by the National Natural Science Foundation of China (No. 62101342, 12326619, 62171290), Guangdong Basic and Applied Basic Research Foundation (No.2023A1515012960); Science and Technology Planning Project of Guangdong Province (No. 2023A0505020002); Shenzhen Science and Technology Program (No. SGDX2020\\1103095613036).

\bibliographystyle{IEEEbib}
\bibliography{strings,refs}

\end{document}